# A New Approach to Lung Image Segmentation using Fuzzy Possibilistic C-Means Algorithm

M.Gomathi
Department of MCA
Velalar College Of Engineering and Technology,
Thindal (PO)
Erode, India
mdgomathi@gmail.com

Dr. P.Thangaraj

Dean, School of Computer Technology and Applications
Kongu Engineering College, Perundurai
Erode, India
ctptr@yahoo.co.in

*Abstract*-Image segmentation is a vital part of image processing. Segmentation has its application widespread in the field of medical images in order to diagnose curious diseases. The same medical images can be segmented manually. But the accuracy of image segmentation using the segmentation algorithms is more when compared with the manual segmentation. In the field of medical diagnosis an extensive diversity of imaging techniques is presently available, such as radiography, computed tomography (CT) and magnetic resonance imaging (MRI). Medical image segmentation is an essential step for most consequent image analysis tasks. Although the original FCM algorithm yields good results for segmenting noise free images, it fails to segment images corrupted by noise, outliers and other imaging artifact. This paper presents an image segmentation approach using Modified Fuzzy C-Means (FCM) algorithm and Fuzzy Possibilistic c-means algorithm (FPCM). This approach is a generalized version of standard Fuzzy C-Means Clustering (FCM) algorithm. The limitation of the conventional FCM technique is eliminated in modifying the standard technique. The Modified FCM algorithm is formulated by modifying the distance measurement of the standard FCM algorithm to permit the labeling of a pixel to be influenced by other pixels and to restrain the noise effect during segmentation. Instead of having one term in the objective function, a second term is included, forcing the membership to be as high as possible without a maximum limit constraint of one. Experiments are conducted on real images to investigate the performance of the proposed modified FCM technique in segmenting the medical images. Standard FCM, Modified FCM, Fuzzy Possibilistic C-Means algorithm (FPCM) are compared to explore the accuracy of our proposed approach.

## I. INTRODUCTION

Image segmentation is a necessary task for image understanding and analysis. A large variety of methods have been proposed in the literature. Image segmentation can be defined as a classification problem where each pixel is assigned to a precise class. Image segmentation is a significant process for successive image analysis tasks.

In general, a segmentation problem involves the division a given image into a number of homogeneous segments, such that the union of any two neighboring segments yields a heterogeneous segment. Numerous segmentation techniques have been proposed earlier in literature. Some of them are histogram based technique, edge based techniques, region based techniques, hybrid methods which combine both the edge based and region based methods together, and so on [1]. In recent years image segmentation has been extensively applied in medical field for diagnosing the diseases.

Image segmentation plays an important role in a variety of applications such as robot vision, object recognition, and medical imaging [2]. In the field of medical diagnosis an extensive diversity of imaging techniques is presently available, such as radiography, computed tomography (CT) and magnetic resonance imaging (MRI) [3, 4]. In recent times, Computed Tomography (CT) is the most effectively used for diagnostic imaging examination for chest diseases such as lung cancer, tuberculosis, pneumonia and pulmonary emphysema. The volume and the size of the medical images are progressively increasing day by day. Therefore it becomes necessary to use computers in facilitating the processing and analyzing of those medical images. Even though the original FCM algorithm yields good results for segmenting noise free images, it fails to segment images corrupted by noise, outliers and other imaging artifact.

Medical image segmentation is an indispensable step for most successive image analysis tasks. This paper presents an image segmentation approach using Modified Fuzzy C-Means (FCM) and Fuzzy Possibilistic C-Means (FPCM) algorithm. Recently, many researchers have brought forward new methods to improve the FCM algorithm [5, 6]. This approach is a generalized version of standard Fuzzy C-Means Clustering (FCM) algorithm. The limitation of the conventional FCM technique is eliminated in modifying the standard technique. The algorithm is formulated by modifying the distance measurement of the standard FCM algorithm to permit the labeling of a pixel to be influenced by other pixels and to restrain the noise effect during segmentation. Possibilistic C-Means (PCM) algorithm, interprets clustering as a Possibilistic partition. Instead of having one term in the objective function, a second term is included, forcing the membership to be as high as possible without a maximum limit constraint of one. Experiments are conducted on real images to investigate the performance of the proposed modified FCM technique in segmenting the medical images. Standard FCM, Modified FCM, Fuzzy Possibilistic C-Means algorithm are compared to explore the accuracy of our proposed approach.

The remainder of the paper is organized as follows. Section 2 provides an overview on related research works in medical image segmentation. Section 3 explains initially explains the standard FCM algorithm and latter it explains the proposed modified FCM and FPCM algorithm. Section 4 discusses on experimental results for real images. Section 5 concludes the paper with fewer discussions.

## II. RELATED WORK

A lot of research work has been carried on various techniques for image segmentation. In recent years, many researchers have brought forward new methods to improve the FCM algorithm [5, 6]. This section of the paper provides an overview on the related research work conducted on medical image processing.

Kenji Suzuki et al. in [7] presented an image processing technique using Massive Training Artificial Neural Networks (MTANN). Their approach resolve the problem faced by radiologists as well as computer-aided diagnostic (CAD) schemes to detect these nodules in case when the lung nodules overlaps with the ribs or clavicles in chest radiographs. An MTANN is extremely a non-linear filter that can be trained by use of input chest radiographs and the equivalent "teaching" images. They used a linear-output back-propagation (BP) algorithm that was derived for the linear-output multilayer ANN model in order to train the MTANN. The dual-energy subtraction is a technique used in [7] for separating bones from soft tissues in chest radiographs by using the energy dependence of the x-ray attenuation by different materials.

A robust statistical estimation and verification framework was proposed by Kazunori Okada et al. in [8] for characterizing the ellipsoidal geometrical structure of pulmonary nodules in the Multi-slice X-ray computed tomography (CT) images. They proposed a multi-scale joint segmentation and model fitting solution which extends the robust mean shift-based analysis to the linear scale-space theory. A quasi-real-time three-dimensional nodule characterization system is developed using this framework and validated with two clinical data sets of thin-section chest CT images. Their proposed framework is a combination of three different but successive stages. They are model estimation, model verification and volumetric measurements. The main issue of the approach is a bias due to the ellipsoidal approximation.

Segmentation-by-registration scheme was put forth by Ingrid Sluimer et al. in [9]. In the scheme a scan with normal lungs is elastically registered to a scan containing pathology. Segmentation-by-registration scheme make use of an elastic registration of inclusive scans using mutual information as a similarity measure. They are compared the performance of four segmentation algorithms namely Refined Segmentation-by-Registration, Segmentation by Rule-Based Region growing, Segmentation by Interactive Region growing, and Segmentation by Voxel Classification. The comparison results revealed that refined registration scheme enjoys the additional benefit since it is independent of a pathological (hand-segmented) training data.

A genetic algorithm for segmentation of medical images was proposed by Ghosh et al. in [10]. In their paper, they presented a genetic algorithm for automating the segmentation of the prostate on two-dimensional slices of pelvic computed tomography (CT) images. In their approach the segmenting curve is represented using a level set function, which is evolved using a genetic algorithm (GA). Shape and textural priors derived from manually segmented images are used to constrain the evolution of the segmenting curve over successive generations. They reviewed some of the existing medical image segmentation techniques. They also compared the results of their algorithm with those of a simple texture extraction algorithm (Laws texture measures) as well as with another GA-based segmentation tool called GENIE. Their preliminary tests on a small population of segmenting contours show promise by converging on the prostate region. They expected that further improvements can be achieved by incorporating spatial relationships between anatomical landmarks, and extending the methodology to three dimensions.

A novel approach for lung nodule detection was described by M. Antonelli et al. in [11]. They described a computer-aided diagnosis (CAD) system for automated detection of pulmonary nodules in computed-tomography (CT) images. Combinations of image processing techniques are used for extraction of pulmonary parenchyma. A region growing method based on 3D geometric features is applied to detect nodules after the extraction of pulmonary parenchyma. Experimental results show, that implementation of this nodule detection method, detects all malignant nodules effectively and a very low false-positive detection rate was achieved.

Xujiong Ye et al. in [12] presented a new computer tomography (CT) lung nodule computer-aided detection (CAD) method. The method can be implemented for detecting both solid nodules and ground-glass opacity (GGO) nodules. Foremost step of the method is to segment the lung region from the CT data using a fuzzy thresholding technique. The next step is the calculation of the volumetric shape index map and the "dot" map. The former mentioned map is based on local Gaussian and mean curvatures, and the later one is based on the Eigen values of a Hessian matrix. They are calculated for each Voxel within the lungs to enhance objects of a specific shape with high spherical elements. The combination of the shape index and "dot" features provides a good structure descriptor for the initial nodule candidate generation. Certain advantages like high detection rate, fast computation, and applicability to different imaging conditions and nodule types make the method more reliable for clinical applications.

A robust medical image segmentation algorithm was put forth by Wang et al. in [13]. Automated segmentation of images has been considered an important intermediate processing task to extract semantic meaning from pixels. In general, the fuzzy c-means approach (FCM) is highly effective for image segmentation. But for the conventional FCM image segmentation algorithm, cluster assignment is based exclusively on the distribution of pixel attributes in the feature space, and the spatial distribution of pixels in an image is not taken into consideration. In their paper, they presented a novel FCM image segmentation scheme by utilizing local contextual information and the high inter-pixel correlation inherent. Firstly, a local spatial similarity measure model is established, and the initial clustering center and initial membership are determined adaptively based on local spatial similarity measure model. Secondly, the fuzzy membership function is modified according to the high inter-pixel correlation inherent. Finally, the image is segmented by using the modified FCM algorithm. Experimental results showed the proposed method achieves competitive segmentation results compared to other FCM-based methods, and is in general faster.

### III. PROPOSED APPROACH

#### A. Conventional Fuzzy C-Means Algorithm

Fuzzy C-Means (FCM) Clustering algorithm is one of the accepted approaches for assigning multi-subset membership values to pixels for either segmentation or other type of image processing [14]. Generally, FCM algorithm proceeds by iterating the two indispensable conditions until a solution is reached. Each data point will be joined with a membership value for each class after FCM clustering. The objective of FCM is to determine the cluster centers and to produce the class membership matrix. In other words, it assigns a class membership to a data point, depending on the similarity of the data point to a scrupulous class relative to all other classes. The class membership matrix is a cXN matrix; in which c is the number of groups and N is the number of samples. Let $X=\{x_1,......x_n\}$ be the training set and $c\geq 2$ be an integer. A fuzzy c-partition of 'X' can be represented by a matrix, U = $\{\mu_{ik}\} \in R^{cXN}$. U can be used to describe the cluster structure of X, by evaluating $\mu_{ik}$, as a degree of membership of $x_k$ to cluster i. The codebook vectors are evaluated by minimizing the distortion measure given by the following equation,

$$\text{Minimize: } J_m(U, v) = \sum_{k=1}^{N} \sum_{i=1}^{c} (\mu_{ki})^m \|X_k - v_i\|^2 A$$

where $X=\{x_1, x_2,...x_N\} \subset R^N$ in a dataset, c is the number of clusters in X: $2 \leq c < N$, m is a weighting exponent: $1 \leq m < \infty$, U = $\{\mu_{ik}\}$ is the fuzzy c-partition of X, $\|X_k - v_i\| A$ is an induced a-norm of $R^N$, and A is a positive-definite (NXN) weight matrix.

A conventional FCM algorithm includes the following steps,

1. Initially values are set for the parameters like c, A, m, ε, and the loop counter 't' is set to 1,
2. As a next step it is necessary to create a random cXN membership matrix U,
3. The cluster centers are then evaluated using the following equation,

$$v_i^{(t)} = \frac{\sum_{k=1}^{N} (\mu_{ki}^{(t)})^m X_k}{\sum_{k=1}^{N} (\mu_{ki}^{(t)})^m}$$

4. The membership matrix is updated periodically with the help of the following equation,

$$\mu_{ki}^{(t+1)} = \left[ \sum_{j=1}^{c} \left( \frac{d_{ki}}{d_{kj}} \right)^{\frac{2}{m-1}} \right]^{-1}$$

Where $d_{ki}$ is given by $\|X_k - v_i^{(t)}\| A$

5. If max $\{|\mu_{ki}^{(t)} - \mu_{ki}^{(t-1)}|\} > \varepsilon$, increment 't' and go to step 3.

#### B. Modified Fuzzy C-Means Clustering Technique for image segmentation

The most important shortcoming of standard FCM algorithm is that the objective function does not think about the spatial dependence therefore it deal with image as the same as separate points. In order to decrease the noise effect during image segmentation, the proposed method incorporates both the local spatial context and the non-local information into the standard FCM cluster algorithm using a novel dissimilarity index in place of the usual distance metric. Therefore a modified FCM algorithm is used to segment the image in our proposed paper. The non-local means algorithm [15] [16] tries to take advantage of the high degree of redundancy in an image. The membership value decides the segmentation results and hence the membership value is evaluated by the distance measurement denoted as $d_{ki}$. Therefore the approach modifies the distance measurement parameter which is readily influenced by local and non-local information.

$$d_{ki}(x_j, v_i) = (1-\lambda_j) d_l^2(x_j, v_i) + \lambda_j d_{nl}^2(x_j, v_i)$$

where $d_l$ stands for the distance measurement influenced by local information, and $d_{nl}$ stands for the distance measurement influenced by non-local information, $\lambda_j$ with the range from zero to one, is the weighting factor controlling the tradeoff between them.

The distance measurement influenced by the local measurement $d_l$ is given by,

$$d_l^2(x_j, v_i) = \frac{\sum_{x_k \in N_j} \omega_l(x_k, x_j) d^2(x_k, v_i)}{\sum_{x_k \in N_i} \omega_l(x_k, x_j)}$$

Where $d^2(x_j, v_i)$ is the Euclidean distance measurement, $\omega_l(x_k, x_j)$ is the weight of each pixel in $N_i$.

The distance measurement influenced by non-local information $d_{nl}$ is computed as a weighted average of all the pixels in the given image I,

$$d_{nl}^2(x_j, v_i) = \sum_{x_k \in I} \omega_{nl}(x_k, x_j) d^2(x_k, v_i)$$

Modified FCM algorithm goes through the following steps,

1. Set the number of clusters 'c' and the index of fuzziness 'm.' Also initialize the fuzzy cluster Centroid vector 'v' randomly and set ε>0 to a small value,
2. Set the neighborhood size and the window size includes the evaluation of cluster centers and membership matrix,
3. Evaluate the modified distance measurement using the equation mentioned as $d_{ki}(x_j,v_i)$,
4. Update the membership matrix and the distance measurement.

### C. Possibilistic C-Means Algorithm (PCM)

In order to overcome the limitations of conventional FCM technique, Possibilistic C-Means (PCM) has been proposed in this paper for medical image segmentation. The Possibilistic C-Means algorithm uses a Possibilistic type of membership function to illustrate the degree of belonging. It is advantageous that the memberships for representative feature points be as high as possible and unrepresentative points have low membership. The intention function, which satisfies the requirements, is formulated as follows,

$$\min \left\{ J_m(x, \mu, c) = \sum_{i=1}^{c} \sum_{j=1}^{N} \mu_{ij}^m d_{ij}^2 + \sum_{i=1}^{c} \eta_i \sum_{j=1}^{N} (1 - \mu_{ij})^m \right\}$$

where, $d_{ij}$ represents the distance between the $j^{th}$ data and the $i^{th}$ cluster center, $\mu_{ij}$ denotes the degree of belonging, m represents the degree of fuzziness, $\eta_i$ is the suitable positive number, c is the number of clusters, and N denotes the number of pixels. $\mu_{ij}$ can be obtained using the following equation,

$$\mu_{ij} = \frac{1}{1 + \left(\frac{d_{ij}^2}{\eta_i}\right)^{\frac{1}{m-1}}}$$

The value of $\eta_i$ determines the distance at which the membership values of a point in a cluster becomes 0.5. The main advantage of this PCM technique is that the value of $\eta_i$ can be fixed or can be changed in each iteration. This can be accomplished by changing the values of $d_{ij}$ and $\mu_{ij}$. The PCM is more robust in the presence of noise, in finding valid clusters, and in giving a robust estimate of the centers.

Updating the membership values depends on the distance measurements [17].The Euclidean and Mahalanobis distance are two common ones. The Euclidean distance works well when a data set is compact or isolated [18] and Mahalanobis distance takes into account the correlation in the data by using the inverse of the variance-covariance matrix of data set which could be defined as follows,

$$D = \sum_{i,j=1}^{i,j=p} A_{ij}(x_i - y_i)(x_j - y_j)$$

$$A_{ij} = \rho_{ij} \sigma_i \sigma_j$$

where, $x_i$ and $y_i$ are the mean values of two different sets of parameters, X and Y. $\sigma_i^2$ are the respective variances, and $\rho_{ij}$ is the coefficient of correlation between $i^{th}$ and $j^{th}$ variants.

### D. Fuzzy Possibilistic C-Means Algorithm (FPCM)

FPCM algorithm was proposed by N.R.Pal, K.Pal, and J.C.Bezdek[18] and it includes both possibility and membership values. FPCM model can be seen as below:

$$\min_{(U,T,V)} \{J_{m,\eta}(U,T,V;X)\} = \sum_{i=1}^{c} \sum_{k=1}^{n} (u_{ik}^m + t_{ik}^\eta) D_{ikA}^2$$

subject to the constraints

$$m > 1, \eta > 1, 0 \leq u_{ik}, t_{ik} \leq 1.$$

$$D_{ikA} = \|x_k - v_i\|_A,$$

and

$$\sum_{i=1}^{c} u_{ik} = 1 \forall k, i.e., U \in M_{fcn}$$

and

$$\sum_{k=1}^{n} t_{ik} = 1 \forall i, i.e., T^t \in M_{fnc}.$$

Where U is membership matrix, T is possibilistic matrix, and V is the resultant cluster centers, c and n are cluster number and data point number respectively. The first order necessary conditions for extreme of $J_{m,\eta}$ are: If $D_{ikA} = \|x_k - v_i\|_A > 0$ for all i and k,m,η > 1 and X contains at least c distinct data points, then

$$(U, T^t, V) \in M_{fcn} \times M_{fcn} \times \mathbb{R}^p$$

may minimize $J_{m,\eta}$ only if

$$u_{ik} = \left(\sum_{j=1}^{c} \left(\frac{D_{ikA}}{D_{jkA}}\right)^{2/(m-1)}\right)^{-1}$$

$$1 \leq i \leq c; 1 \leq k \leq n$$

$$t_{ik} = \left(\sum_{j=1}^{n}\left(\frac{D_{ikA}}{D_{ijA}}\right)^{2/(\eta-1)}\right)^{-1}$$
$$1 \leq i \leq c; 1 \leq k \leq n$$
$$v_i = \frac{\sum_{k=1}^{n}(u_{ik}^m + t_{ik}^\eta)x_k}{\sum_{k=1}^{n}(u_{ik}^m + t_{ik}^\eta)}, 1 \leq i \leq c.$$

The above equations show that membership $u_{ik}$ is affected by all c cluster centers, while possibility $t_{ik}$ is affected only by the i-th cluster center $c_i$. The possibilistic term distributes the $t_{ik}$ with respect to all n data points, but not with respect to all c clusters. So, membership can be called relative typicality, it measures the degree to which a point belongs to one cluster relative to other clusters and is used to crisply label a data point. And possibility can be viewed as absolute typicality, it measures the degree to which a point belongs to one cluster relative to all other data points, it can reduce the effect of outliers. Combining both membership and possibility can lead to better clustering result.

## IV. EXPERIMENTAL RESULTS

The proposed Modified FCM algorithm and Fuzzy Possibilistic C-Means algorithm is implemented using MATLAB and tested on real images to explore the segmentation accuracy of the proposed approach. Three various types of FCM techniques has been used in the experiments for comparison. They are standard FCM, Modified FCM, and Fuzzy Possibilistic C-Means Clustering algorithm.

### A. Real Image Dataset

A real set of lung images are used to evaluate the accuracy of the proposed algorithm in segmenting the medical images. The results obtained are then compared with the segmentation results that were performed manually to explore the accuracy of the proposed algorithm. The segmentation results of standard FCM, Modified FCM, and Fuzzy Possibilistic C-Means Clustering are considered to investigate the best algorithm that delivered better segmentation results for real medical images. The three most important parameters used to determine the accuracy of the proposed algorithm are similarity, false positive and the false negative ratio. From the results obtained it can be concluded that our proposed algorithm performed well head of other techniques in segmenting the real medical images. The three main attributes mentioned above i.e. similarity, false positive ratio, and the false negative ratios are listed in Table 1, for all the three image segmentation techniques. Figure 1 shows the segmentation result of standard FCM and FPCM.

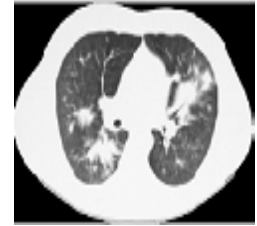

*(a)*

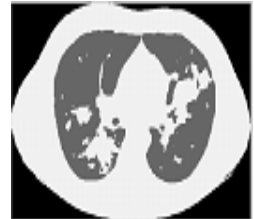 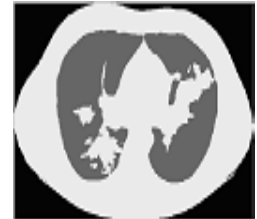

*(b)* *(c)*

Figure 1 (a) Actual Image, Segmented Images (b) using Standard FCM (c) using FPCM

Table 1 Different Indices for Different Algorithms

| Algorithm | Similarity | False Positive Ratio | False Negative Ratio |
|---|---|---|---|
| Standard FCM | 86.03 | 20.15 | 8.50 |
| Modified FCM | 89.50 | 16.50 | 5.30 |
| Fuzzy Possibilistic C-Means Clustering (FPCM) | 92.50 | 12.80 | 3.40 |

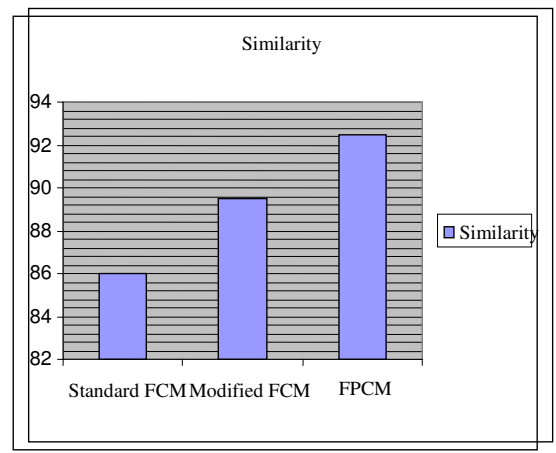

Figure 2 Comparison of Similarity

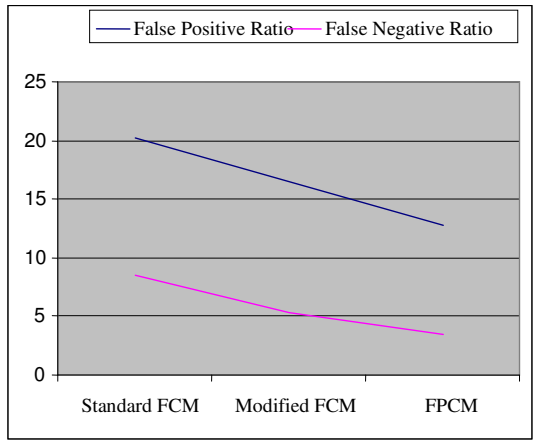

Figure 3 Comparison of False Positive and False Negative Ratio for the three approaches

The experimental results obtained by employing the Fuzzy Possibilistic C-Means (FPCM) algorithm revealed that the proposed technique of image segmentation has a better performance over other FCM extension methods. Furthermore, the proposed approach of image segmentation using Fuzzy Possibilistic C-Means algorithm eliminates the effect of noise greatly. This in turn increased the segmentation accuracy of the proposed image segmentation technique.

## V. CONCLUSION

FCM is one of an accepted clustering method and has been broadly applied for medical image segmentation. Conversely, conventional FCM at all times suffers from noise in the images. Even though the original FCM algorithm yields good results for segmenting noise free images, it fails to segment images corrupted by noise, outliers and other imaging artifact. Although many researchers have developed a variety of extended algorithms based on FCM, not any of them are perfect. A modified FCM clustering algorithm and Fuzzy Possibilistic C-Means (FPCM) algorithm is proposed in this paper. In the proposed Modified FCM algorithm, both local and non-local information are incorporated to control the tradeoff between them. The algorithm is formulated by modifying the distance measurement of the standard FCM algorithm to permit the labeling of a pixel to be influenced by other pixels and to restrain the noise effect during segmentation. The Fuzzy Possibilistic C-Means (FPCM) algorithm interprets clustering as a Possibilistic partition and includes membership functions. Instead of having one term in the objective function, a second term is included, forcing the membership to be as high as possible without a maximum limit constraint of one. Experiments are conducted on real medical images to evaluate the performance of the proposed algorithm. The three most important parameters used to determine the accuracy of the proposed algorithm are similarity, false positive and the false negative ratio. The experimental results suggested that the proposed algorithm performed well than other FCM extension, segmentation algorithms.

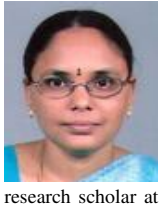
**M. Gomathi** received the Master of Computer Applications degree in 1998 from Bharathiar University. Her Masters Thesis research was on "Handwritten Character Recognition", and has completed her M. Phil, in Computer Science at Bharathiar University in the year 2004. Currently she is working as an Assistant Professor in Velalar College of Engineering and Technology and a research scholar at Kongu Engineering College, Anna University. Her area of research is in Medical Imaging.

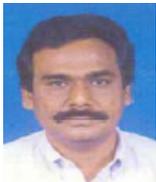
**Dr. P. Thangaraj** received the Bachelor of Science degree in Mathematics from Madras University in 1981 and his Master of Science degree in Mathematics from the Madras University in 1983. He completed his M. Phil degree in the year 1993 from Bharathiar University. He completed his research work on Fuzzy Metric Spaces and awarded Ph. D degree by Bharathiar University in the year 2004. He completed the post graduation in Computer Applications at IGNOU in 2005. His thesis was on "Efficient search tool for job portals". He completed his Master of Engineering degree in Computer Science in the year 2007 from Vinayaka Missions University. His thesis was on "Congestion control mechanism for wired networks". Currently he is designated as Dean of School of Computer Technology and Applications at Kongu Engineering College, Autonomous institution affiliated to Anna University. His current area of research interests are in Fuzzy based routing techniques in Ad-hoc Networks.